\documentclass[sigconf, 9pt, screen]{acmart}

\usepackage{enumitem}
\usepackage{tikz}
\usetikzlibrary{arrows.meta, positioning}
\usepackage[para]{footmisc}
\usepackage{tabularx}

\AtBeginDocument{%
  }

\copyrightyear{2026}
\acmYear{2026}
\setcopyright{cc}
\setcctype{by}
\acmConference[ACM DXConf 2026]{The Inaugural ACM Conference on Digital Transformation}{October 29--30, 2026}{Ann Arbor, MI, USA}
\acmBooktitle{The Inaugural ACM Conference on Digital Transformation (ACM DXConf 2026), October 29--30, 2026, Ann Arbor, MI, USA}
\acmDOI{10.1145/3847238.3849037}
\acmISBN{979-8-4007-3028-3/2026/10}

\begin{document}

\title[A Global Comparison of AI Registers]{A Global Comparison of Schemas, Transparency, and Interoperability in Public-Sector AI Registers and Inventories}

\author{Dipto Das}
\authornote{The author worked on this project while working at the University of Toronto.}
\affiliation{
  \institution{Cornell University}
  \city{Ithaca}
  \state{New York}
  \country{United States}
}
\email{dd749@cornell.edu}

\author{Shion Guha}
\affiliation{
  \institution{University of Toronto}
  \city{Toronto}
  \state{Ontario}
  \country{Canada}
}
\email{shion.guha@utoronto.ca}

\ccsdesc[500]{Social and professional topics~Governmental regulations}
\ccsdesc[500]{Applied computing~E-government}

\keywords{AI, Register, Inventory, Schema, Layered Visibility Framework}

\begin{abstract}
  Artificial intelligence (AI) registers and inventories aim to make governmental AI visible, but their institutional scope, schemas, and reporting practices construct different representations of public-sector AI. We compare 8,368 records from country-specific and transnational inventories covering 72 countries. Across 23 harmonized fields, registers shared a descriptive core but rarely requested information about appeals, risks, legal bases, or external evaluation. We found that broad schemas often contained substantial missingness, schema similarity showed no significant patterned convergence, and multiple sources covering the same jurisdictions overlapped only selectively. Based on these findings, we synthesize a \emph{layered visibility framework} that shows how register records reflect disclosure arrangements and why interoperability requires shared concepts, clear definitions, and preserved provenance.
\end{abstract}

\maketitle
\section{Introduction}\label{sec:introduction}
While AI is used in at least one area of government in 35 of 36 countries surveyed by the Organization for Economic Co-operation and Development, only three reported mandatory AI use-case repositories, while another ten maintained repositories based on optional agency contributions~\cite{oecd2026digitalgovernment}. This disparity is increasingly consequential as governments use AI to transform public services, internal operations, policymaking, and administrative decision-making, while its capabilities advance faster than the evidence and institutional safeguards needed to govern them~\cite{unpanel2026preliminary}. Thus, governments face a fundamental digital-transformation problem: they deploy AI faster than they build shared infrastructure to identify, document, and scrutinize it. Public AI inventories and registers have emerged in response, but their institutional mandates, schemas, reporting units, and collection processes differ substantially. We refer to these collectively as AI-inventorying instruments. This paper compares how these instruments are designed and what representations of governmental AI they produce.

AI registers extend structured disclosure from individual systems to portfolios used across public institutions. Although initiatives have emerged at municipal~\cite{murad2021beyond,haataja2020public}, national~\cite{das2026bureaucratic}, and transnational~\cite{pi2026understanding} levels, prior research has generally examined them individually. Consequently, we know comparatively little about how their schemas and reporting practices differ or whether sources covering the same jurisdiction enumerate the same systems. Rather than treating registers as interchangeable databases or censuses of governmental AI, we compare them as policy instruments whose reporting authorities, inclusion rules, schemas, classifications, and collection procedures shape what becomes publicly visible and governable~\cite{lascoumes2007introduction}. Because policy models can circulate through learning, emulation, and transnational networks while being adapted to local conditions~\cite{dobbin2007global,marsh2013policy,stone2004transfer}, we also examine whether schema similarities exhibit temporal or geographic clustering. We ask: \textbf{RQ1.} How do AI-inventorying instruments differ in schema breadth and field completion, and do their schemas exhibit temporal convergence or geographic clustering? \textbf{RQ2.} How do the portfolios represented by different AI-inventorying instruments vary in lifecycle composition and organizational concentration? \textbf{RQ3.} To what extent do country-specific and transnational instruments covering the same jurisdiction report the same AI systems?

We conducted a comparative quantitative study of 8,368 records from 17 country-specific and transnational instruments covering 72 countries, along with the regional and global-level records. The instruments shared a descriptive core but showed no evidence of temporal convergence or clear geographic clustering. Broader schemas were not necessarily more complete, represented portfolios differed in lifecycle composition and organizational concentration, and sources covering the same jurisdictions overlapped only selectively. These findings show that AI registers do not directly mirror governmental AI activity but construct visibility through institutional scope, schema design, and reporting implementation. Drawing these relationships together, we develop a \textbf{\emph{layered visibility framework}} comprising institutional scope, schema affordances, reporting realization, and public representation. The framework treats visibility not as a binary property of whether a register exists, but as an outcome of decisions about which systems enter it, what its schema can represent, whether available fields are populated, and what portfolio consequently becomes visible. It provides a diagnostic vocabulary for distinguishing limitations arising from institutional coverage, schema design, reporting practice, and the resulting representation of governmental AI landscape.
\section{Literature Review}\label{sec:literature_review}
We situate AI registers within research on digital transformation and public-sector governance, documentation infrastructures, and policy diffusion, convergence, and interoperability. Together, these perspectives explain how institutional conditions, schemas, and reporting practices shape the public visibility of governmental AI.

\subsection{AI Governance in Digital Transformation}
Digital transformation involves more than adopting new technologies or converting existing information into digital formats. It entails broader changes to organizational structures, institutional relationships, and value-creation processes~\cite{vial2021understanding}. Within government, it similarly involves redesigning administrative processes and public services to pursue objectives such as efficiency, interoperability, transparency, and citizen responsiveness~\cite{mergel2019defining}. Public-sector AI adoption is one component of this transformation: as agencies introduce computational systems to automate tasks, it also reorganizes how information is collected, decisions are supported, services are delivered, and responsibilities are distributed among officials, technical systems, and external providers~\cite{saxena2021framework}.

Public-sector AI governance is commonly articulated through principles such as transparency, accountability, fairness, safety, privacy, and trustworthiness~\cite{jobin2019global,diaz2023connecting,felzmann2019transparency}. However, bureaucratic organizations with different missions, legal responsibilities, resources, procurement arrangements, and technical capacities must operationalize these principles~\cite{gagua2025responsible,saxena2021framework,veale2018fairness}. Algorithmic accountability is therefore not solely a property of system design but an institutional practice shaped by governing rules, professional judgment, organizational incentives, and relationships among agencies, technology providers, workers, and affected publics~\cite{binns2018algorithmic,wieringa2020account,johnson2021algorithmic,moon2025datafication}. It depends partly on whether institutions identify the systems they use, document their roles in administrative processes, trace the actors involved, and disclose this information to relevant publics~\cite{kroll2021outlining}.

Even when information is public, transparency does not necessarily produce accountability or make a system understandable, actionable, or contestable~\cite{ananny2018seeing,kizilcec2016much}. Transparency is better understood as a situated communicative arrangement in which the content, format, audience, and institutional conditions of disclosure shape what recipients can know and do~\cite{eyert2023rethinking,norval2022disclosure}. From this perspective, public AI inventories are not merely websites or datasets but administrative mechanisms that determine which systems are reported, how they are classified, who maintains their records, and which aspects of public-sector AI become externally visible.

\subsection{AI Registers as Disclosure Infrastructures}
Documentation frameworks translate abstract commitments to transparency and accountability into standardized, inspectable artifacts. Datasheets~\cite{gebru2021datasheets}, model cards~\cite{mitchell2019model}, audit reports~\cite{raji2020closing}, impact assessments~\cite{watkins2021governing}, data statements~\cite{mcmillan2024data}, and other reporting standards structure disclosures about provenance, intended use, limitations, risks, performance, and responsible actors. These artifacts can support coordination among developers, deploying organizations, regulators, auditors, and affected communities~\cite{vertesi2026reckoning,moon2025datafication,das2026asymmetries}. Simultaneously, documentation practices reflect institutional norms and assumptions about what information is relevant, which audiences matter, and what forms of responsibility should be recorded~\cite{dergacheva2023one,poirier2022accountable,boag2022tech}. Therefore, documentation instruments such as AI inventories shape which system properties become legible, which actors are rendered responsible, and which accountability claims can be made~\cite{poirier2021reading,kuehnert2025and}.

Critical data studies scholarship has shown how claims of scale, neutrality, completeness, and representativeness can obscure the social and organizational conditions under which datasets and systems are produced~\cite{kitchin2016makes,denton2021genealogy}. Standardized documentation can similarly reduce complex concerns about fairness and accountability to questions of data sufficiency or technical performance while making data labor, contextual knowledge, and organizational practice less visible~\cite{aragon2022human,groesch2025big,weerts2024neutrality}. Thus, documentation can enable scrutiny while simultaneously delimiting its scope.

AI registers and inventories extend documentation from individual datasets or models to portfolios of algorithmic systems used across organizations. Municipal governments such as Amsterdam and Helsinki were among the earliest adopters of public algorithm registers, publishing structured information about systems used in public services~\cite{murad2021beyond,haataja2020public,sieber2026building}. Similar initiatives subsequently appeared in Nantes, Antibes, Lyon, and other municipalities, as well as at national and transnational levels across North America and Europe~\cite{ogp2021building,kaushal2024automated}. Registers also function as institutional ontological design~\cite{das2026bureaucratic}: their categories define which systems and properties become administratively legible. Producing them depends on often-invisible work, including identifying relevant systems, interpreting rules, coordinating across organizations, translating contextual practices into standardized fields, validating submissions, and updating records as systems move through their lifecycles~\cite{poirier2022accountable,chappidi2025accountability}.

Hence, it is important to distinguish between what an instrument can structurally represent and the extent to which organizations populate its available fields. For example, a register that requests information about procurement, human involvement, or appeal mechanisms creates different accountability possibilities than one limited to system names and short descriptions. Disclosure templates, reporting thresholds, machine-readable categories, and organizational incentives and interpretations shape both what institutions report and what remains difficult to disclose~\cite{nieuwenhuizen2024algorithm,hogberg2024stabilizing,kaushal2024automated}. Transparency databases can consequently standardize reporting while still providing limited contextual explanation or falling short of their regulatory purposes~\cite{trujillo2025dsa,drolsbach2024content,groesch2025big,poirier2022accountable}. Because instruments differ in institutional coverage, reporting obligations, and how they treat planned, operational, and retired systems, the portfolios they disclose may also differ in lifecycle composition and organizational concentration. We should interpret these distributions as properties of particular disclosure arrangements rather than as direct measures of governmental AI adoption. A public search interface or database-like inventory can be understood as the visible outcome of broader coordination among policy requirements, administrative labor, data schemas, and public communication~\cite{nieuwenhuizen2024algorithm,winecoff2025improving}.

\subsection{Policy Diffusion and Interoperability}
AI registers can also be understood as policy instruments. Policy scholarship defines instruments as technical and social devices that organize relationships between governing institutions and those they seek to govern~\cite{lascoumes2007introduction}. Because of their sociopolitical embeddedness, public AI registers and inventories should not be treated as neutral. Their reporting authorities, inclusion criteria, schemas, submission and update procedures, and public interfaces privilege particular forms of knowledge, assign responsibilities, and structure administrative behavior. As a result, instruments that share the general objective of making governmental AI visible may embody substantially different accountability models.

The growing adoption of AI registers worldwide also raises questions of diffusion and transfer: how policy spreads through learning, competition, coercion, and the construction of shared norms by professional and transnational communities~\cite{dobbin2007global}. International organizations, professional networks, civil-society initiatives, consultants, and other non-state actors can serve as transfer agents by circulating policy knowledge and facilitating the uptake of ideas, administrative arrangements, formal instruments, and informal norms of appropriate governance across various settings~\cite{marsh2013policy,stone2004transfer}.

However, diffusion does not necessarily produce uniformity. Jurisdictions may copy an existing template, emulate selected features, combine elements from several models, or reinterpret a shared idea in light of local administrative arrangements~\cite{jarvers2026engaged}. For example, similar labels such as ``AI register," ``algorithm dashboard," or ``use-case inventory" can consequently refer to instruments with different reporting units, legal statuses, collection procedures, or public functions~\cite{pi2026understanding}. This makes schema interoperability---the capacity to represent and interpret corresponding concepts across heterogeneous instruments---a central concern. Instruments developed independently may nevertheless converge around a descriptive core as they confront similar administrative problems. Temporal proximity and participation in the same institutional networks may facilitate such convergence~\cite{elkins2005waves,maggetti2016problems}, though particular diffusion mechanisms can vary. Observed similarity may also arise from common technical requirements, widely circulating governance norms, or the limited set of categories readily available for describing AI systems~\cite{holzinger2005causes, dimaggio1983iron}.

Prior work has largely examined individual municipal, national, or jurisdictional initiatives, particularly in Western and Global North contexts~\cite{cath2021dutch,kingsman2022public,murad2021beyond,das2026bureaucratic}. We know comparatively less about how schemas and reporting completion vary across instruments, whether schema similarities exhibit temporal, geographic, or institutional patterns, and how the portfolios represented by different instruments vary in composition. For jurisdictions covered by both country-specific and transnational repositories, it also remains unclear whether these sources identify the same systems or construct different representations of governmental AI. In particular, identifying a diffusion mechanism requires evidence of relationships among adopters, such as direct template borrowing, technical assistance, professional-network participation, or policymakers' design decisions. Moreover, cross-jurisdictional comparison requires sufficient conceptual agreement about what fields represent and how their categories are defined, while preserving differences in provenance and context. Such comparison requires distinguishing three analytical units. An \emph{instrument} is a sociotechnical device with its own institutional purpose and reporting process~\cite{lascoumes2007introduction}. In this study, an instrument may be a register, inventory, or transnational repository. A \emph{schema} specifies the concepts, attributes, and relationships through which information can be represented~\cite{chen1976entity}. The temporal and geographic clustering of schemas and instruments can be treated as exploratory indicators of diffusion-consistent convergence rather than tests of learning, competition, coercion, or normative emulation. Observed similarity may result from diffusion, but it may also arise because instruments confront similar administrative problems, draw on widely circulating governance norms, or rely on a limited vocabulary for describing AI systems. A \emph{source record} is an individual structured entry containing values for the fields defined by that schema~\cite{codd1970relational}, which was produced through a particular instrument's collection process. These distinctions let us compare schema breadth separately from record-level completion.
\section{Methods}
Our analysis proceeded across three levels corresponding to the research questions. First, we compared instruments' schema breadth and field completion and examined temporal and geographic patterns in schema similarity. Second, we compared the lifecycle composition and organizational concentration of the portfolios represented by each instrument. Third, for jurisdictions covered by both country-specific and transnational sources, we identified corresponding records and compared their overlap, disclosure completion, and lifecycle composition.

\subsection{Source Identification and Eligibility}
We searched combinations of each country's name\footnote{\url{www.un.org/en/about-us/member-states}} and terms, such as \emph{AI register}, \emph{AI inventory}, \emph{algorithm register}, and \emph{automated decision system inventory}, between June 26, 2026 and July 11, 2026. We included resources that (1) operated at the national or federal level, sought countrywide coverage, or cataloged multiple jurisdictions through a common transnational framework; (2) enumerated identifiable AI systems, algorithms, projects, or use cases; and (3) were publicly accessible or publicly documented. We excluded city- or agency-specific registers, general AI policy documents, and aggregated statistics and reports. However, countrywide and transnational sources could include local or agency-level records.

\subsection{Data Provenance and Corpus Construction}
\begin{figure}[!ht]
    \centering
    \includegraphics[width=\linewidth, height=1.5in]{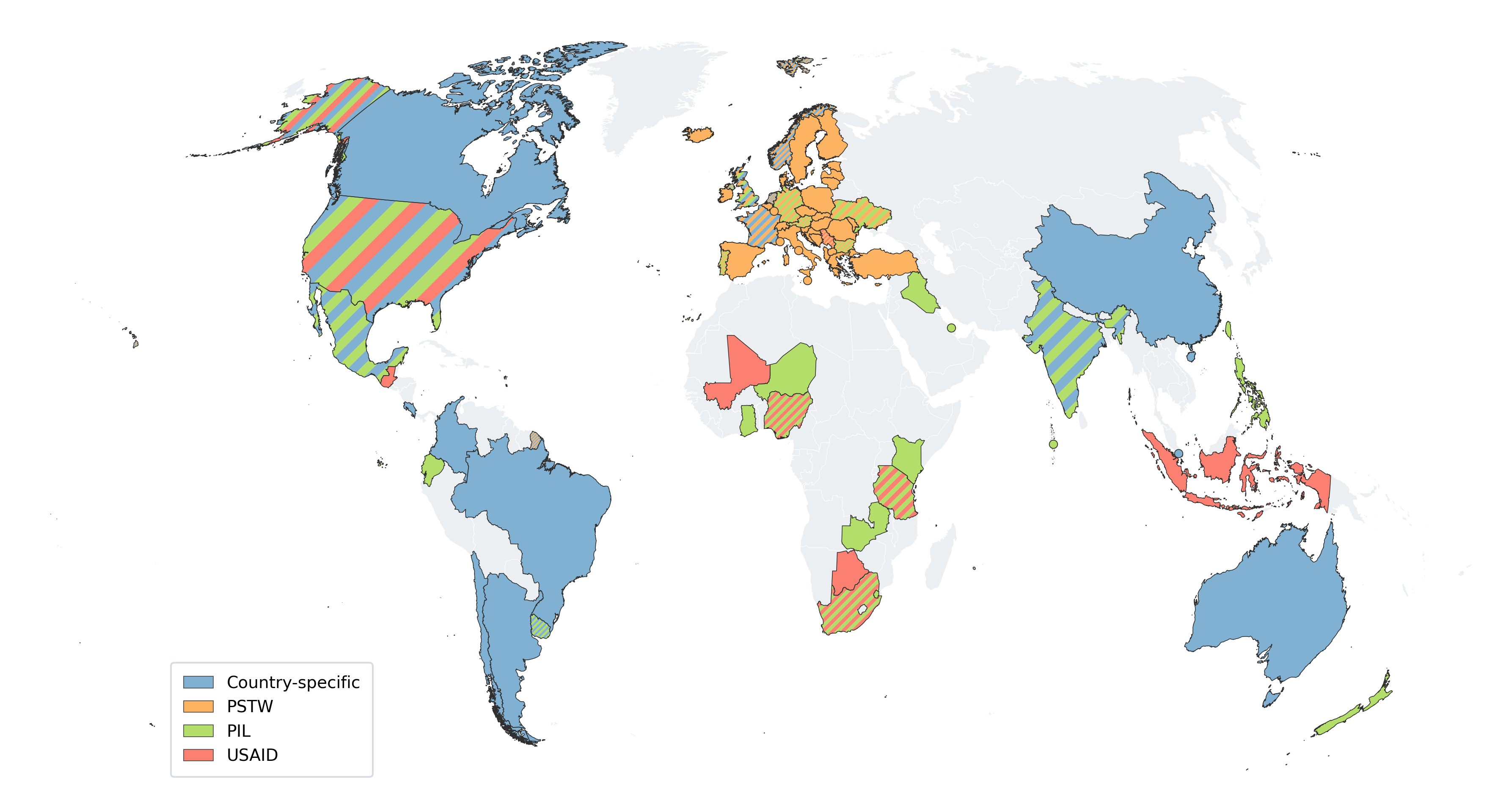}
    \caption{Geographic coverage of AI register sources. Diagonal stripes indicate countries covered by multiple sources.}
    \label{fig:choropleth}
\end{figure}

Our provenance audit identified two principal source families. We identified 17 country-specific sources from Argentina, Australia, Brazil, Canada, Chile, China, Colombia, Costa Rica, France, India, Mexico, the Netherlands, Norway, Singapore, the UK, the US, and Uruguay. Australia's use case library\footnote{\url{www.govai.gov.au/explore/use-cases}} requires special membership, and China's Beian\footnote{\url{beian.miit.gov.cn/Integrated/index}} and India's AIKosh\footnote{\url{aikosh.indiaai.gov.in/home/use-cases/all}} provided public lookup interfaces, but we could not identify an official release of their underlying datasets or APIs and hence excluded these three as standalone sources. For several other countries without directly downloadable datasets, including Argentina, Brazil, and Norway, we converted publicly available RSS\footnote{Really Simple Syndication: an automated XML web feed for content distribution.} or Omeka feeds and HTML pages into tabular form. The second family consisted of repositories compiled by the Public Sector Tech Watch (PSTW), the Policy Innovation Lab (PIL), and the US Agency for International Development (USAID)\footnote{PIL and USAID inventories retrieved from the same source were treated as combined.}. From these repositories, we extracted records covering 65 countries, along with EU-level and global records, where AI was identified as the primary technology. Across both source families, the corpus contained information about AI use in 72 countries, along with EU-level and global records (Figure~\ref{fig:choropleth}). Some countries, including France, Mexico, and the US, appeared in both families. Although we excluded AIKosh as a standalone source, we retained records of Indian AI applications found in the transnational repositories. Following prior work~\cite{wang2023document, rehlinger2026fine}, we used GPT-5 to translate non-English documents into English.

Because the sources varied in both schema and reporting unit--describing systems, use cases, or applications--we treated each row as an observation produced by a particular inventorying process rather than as a directly comparable measure of AI adoption. We retained the original source fields while constructing a harmonized analytical layer of 23 conceptually comparable fields. These fields covered provenance and identification; system description, responsible organization, lifecycle, administrative function, and technology; intended users and development responsibility; data, procurement, and human involvement; legal basis, risk, and evaluation; source-code availability; and public contact mechanisms.

We mapped fields at the conceptual rather than literal-name level. A source could contain multiple columns corresponding to one harmonized concept. For example, we consolidated separate questions and explanatory fields concerning procurement, human involvement, or risk. Similarly, \emph{``human review required"}, \emph{``human intervention"}, and \emph{``human involved in the final decision"} were mapped to a shared human-involvement field while preserving their source-specific wording and definitions. Conversely, we excluded administrative and technical metadata and other source-specific fields. Thus, a country's or transnational entity's schema breadth indicates the number of harmonized concepts it represented, not its total number of columns in the dataset originally released by them\footnote{For example, the US AI Inventory dataset contains 36 fields, some of which were consolidated into broader concepts or excluded from the harmonized comparison.}.

We harmonized administrative-function and technology categories only when correspondence between source categories was sufficiently direct. Where possible, we likewise mapped lifecycle values into planned, in development, pilot, operational, completed, retired, or unknown. We did not impute absent information and excluded ambiguous values from cross-source comparisons.

The resulting corpus contained 8,368 records, which did not necessarily represent unique systems because the same system could appear in multiple sources. We assigned each record a provenance key combining its source instrument, original identifier, and reporting unit. To analyze missingness, we distinguished \emph{structural omission}, where an instrument did not request a field, from \emph{reporting omission}, where the field existed but was unpopulated. This distinction separated \emph{structural visibility}, the fields a register made available for disclosure, from \emph{reporting realization}, the extent to which individual records completed those fields.

\subsection{Comparative Data Analysis}

\subsubsection{Register Design and Disclosure}
We represented each instrument as a 23-field binary vector indicating whether each harmonized field was present in its schema. We calculated pairwise Jaccard similarity and used average-linkage agglomerative clustering based on Jaccard distance to order an exploratory register-by-field heatmap. We included adoption date and geographic region as annotations to aid interpretation. To examine diffusion-consistent convergence, we used node-label permutation tests~\cite{krackhardt1988predicting}, which preserve the dependence structure created by instrument pairs sharing the same nodes. The temporal test compared mean Jaccard similarity between pairs adopted within two years of one another and pairs adopted more than two years apart. For each test, we permuted the corresponding adoption year across instrument nodes, recalculated the difference in group means, and used a one-sided test of whether temporally proximate pairs exhibited greater schema similarity.

For each field present in an instrument's schema, we calculated completion as the proportion of eligible records containing a reported value. We visualized schema presence and field completion as separate register-by-field matrices, distinguishing narrow schemas, broad and consistently populated schemas, and broad schemas with substantial reporting omission.

\subsubsection{Lifecycle Composition and Organizational Concentration}
Because governmental level, institutional sector, administrative function, and technology type were represented using highly source-specific vocabularies, cross-instrument portfolio analysis focused on the harmonized lifecycle variable and within-instrument organizational concentration. We calculated record-weighted lifecycle totals to describe the assembled corpus. We then compared the lifecycle distributions of instruments with usable harmonized lifecycle information using pairwise Jensen--Shannon divergence (JSD)~\cite{lin1991divergence}. JSD ranges from zero for identical distributions to one for distributions with no overlap. To examine whether disclosed records were broadly distributed or concentrated among a small number of reporting organizations, we calculated the normalized Herfindahl--Hirschman Index (HHI) for each instrument~\cite{cracau2016normalized}. The normalized HHI measure ranges from zero for an equal distribution to one for complete concentration. We also reported the shares contributed by the largest and five largest organizations to aid interpretation.


\subsubsection{Cross-Source Representation}
For jurisdictions represented in both country-specific and transnational sources, we generated potentially corresponding record pairs using normalized system titles, responsible-organization names, URLs, dates, and semantic similarity between descriptions. This candidate-generation procedure produced 490 pairs, which we manually reviewed and confirmed 81 one-to-one matches. We accepted a candidate only when comparisons of the available identifying fields indicated that both records described the same underlying system or use case; we did not treat ambiguous candidates as matches. We classified records without a verified counterpart as source-unique for that source pair, which does not establish that the underlying system was absent from the other source. For each source pair, we calculated the number of verified shared records, the number and proportion of records unique to each source, Jaccard overlap, and directional coverage. Then, we compared disclosure completion and lifecycle composition between shared and source-unique records.


\subsection{Limitations}
Cross-source overlap should be interpreted as verified, not exhaustive. Candidate generation depended on the identifying information available in each source, and no complete ground truth existed for estimating recall. Differences in language, naming, update timing, reporting units, and aggregation may have prevented us from identifying genuine counterparts. Therefore, in the context of our paper, ``source-unique" means lacking a verified counterpart, not that the underlying system was definitively absent from the other source.
\section{Results}
We present the results in three parts. First, we compare schema breadth, field completion, and patterns of schema similarity. Second, we examine differences in lifecycle composition and organizational concentration across instruments. Third, we assess the overlap between country-specific and transnational sources and compare the records they share with those unique to each source.

\subsection{Register Design and Schema Similarity}\label{sec:results-schema}
Across all instruments, we identified 23 distinct disclosure fields. The schemas shared a basic descriptive core: every instrument disclosed a system name and responsible organization, 15 identified the technology type, and 14 included a description and lifecycle status. Fields related to contestability and external accountability were considerably less common. Only one instrument (the US) included an appeal mechanism, while two instruments included risks (the Netherlands and the US), legal basis (France and the Netherlands), and external evaluation (France and the US).

\begin{figure*}[!ht]
    \centering
    \includegraphics[width=0.81\textwidth]{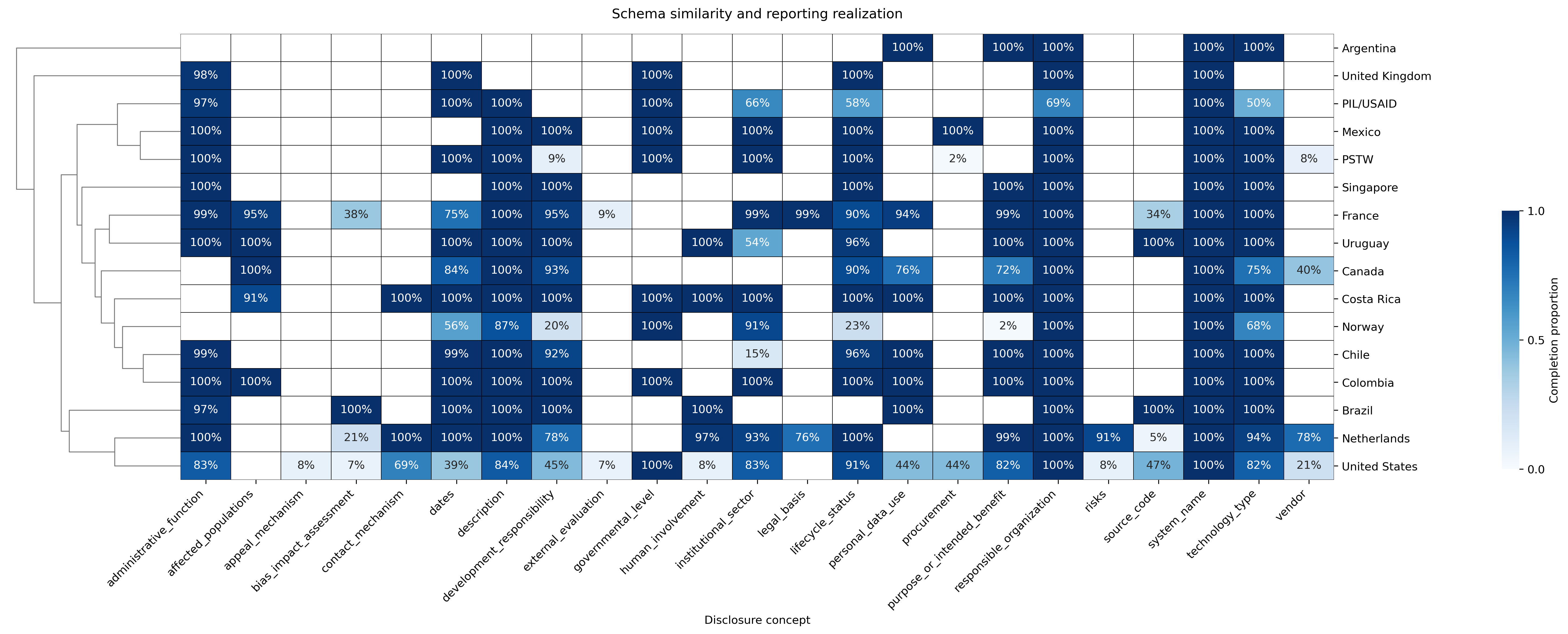}
    \caption{Schema similarity and record-level completion across AI inventories. Blank cells indicate fields' absence from schemas.}
    \label{fig:schema-completion}
\end{figure*}

Across the 120 unique instrument pairs, Jaccard schema similarity ranged from $0.21$ to $0.85$ (mean $J=0.53$). The most similar pair, Chile and Colombia, represented two Latin American countries. However, the second-most similar pair, Mexico--PSTW ($J=0.83$), crossed geographic regions and source types. Hierarchical ordering revealed no clear geographic groupings (see Figure~\ref{fig:schema-completion}).

Schema presence did not necessarily mean that the corresponding information was consistently reported. Across 187 eligible schema--field combinations, 115 (61.5\%) were complete for every eligible record, whereas 24 (12.8\%) had completion rates below 50\%. Mean completion across available fields ranged from 54.9\% in the US inventory to 100\% in the Argentina, Colombia, Mexico, and Singapore inventories. These comparisons must be considered alongside schema breadth: the US inventory contained 21 disclosure fields, whereas the Argentina and Singapore inventories contained only five and eight, respectively. Thus, broad schemas did not necessarily produce more complete disclosure.

The 82 instrument pairs adopted within two years of one another had a mean similarity of $0.540$, compared with $0.504$ among the 38 pairs adopted more than two years apart. This difference was small and not statistically significant (mean difference $=0.036$, one-sided permutation $p=0.184$), providing no evidence that temporal proximity was associated with greater schema similarity.

\subsection{Lifecycle Divergence and Organizational Concentration}\label{sec:results-portfolios}
Across instruments, responsible organization and lifecycle stage were reported for 98.06\% and 88.85\% of records, respectively. Among records with a reported lifecycle stage, operational systems constituted the largest category (49.5\%), followed by systems in development (26.9\%), pilots (16.6\%), retired systems (6.5\%), and planned systems (0.3\%). Lifecycle distributions varied sharply across instruments. Operational systems constituted all 10 usable records in Costa Rica, 97.5\% of records in Mexico, and 91\% in the Netherlands. In the second source family, 79\% and 42.56\% of systems reported by PIL/USAID and PSTW, respectively, were in use. Colombia presented a contrasting pattern, with 65.5\% of its records classified as in development. Systems in development also constituted 50\% of Uruguay's, 44\% of Canada's, and 44.6\% of the US inventory. Pilots were especially prominent in PSTW (40.9\%).



Across 14 eligible instruments and 91 instrument pairs, lifecycle JSD ranged from $0.013$ between Costa Rica and Mexico to $0.845$ between Colombia and Costa Rica, with a mean of $0.250$. These two comparisons should be interpreted cautiously because Costa Rica contributed only 10 usable lifecycle records. Other large divergences involved Colombia's development-heavy portfolio, including Colombia--Mexico ($0.798$), Colombia--Chile ($0.711$), and Colombia--Netherlands ($0.698$). Among the most similar larger portfolios were France and the UK ($0.025$) and the UK and PSTW ($0.038$).

Organizational concentration also varied substantially. Normalized HHI values ranged from $0$ to $0.17$ (median $=0.018$). Singapore's technology-offer portfolio had the highest concentration ($HHI=0.17$): the National University of Singapore accounted for 46.7\% of its records, and the five most represented organizations collectively accounted for 93.3\%. Uruguay ($0.07$), PIL/USAID ($0.06$), and the US ($0.05$) exhibited the next-highest values. In the US inventory, the largest organization, the National Aeronautics and Space Administration, contributed 14.6\% of records, while the five largest organizations collectively accounted for 58.2\%. In contrast, PSTW contained records from 1,415 organizations in the EU region and had a normalized HHI of $0.0006$. Its largest organization, the Portuguese Agency for Administrative Modernization, accounted for only 1.1\% of records, and its five largest organizations together accounted for 4.2\%. Mexico ($0.002$), Chile ($0.003$), and the Netherlands ($0.008$) also exhibited comparatively dispersed portfolios.

\subsection{Cross-Source Representation and Overlap}\label{sec:results-overlap}
Cross-source comparison was possible for eight pairs across seven countries and the two transnational repositories, PSTW and PIL/USAID. Norway exhibited the greatest overlap with PSTW: 48 records appeared in both sources (Jaccard overlap $=0.245$). Although 73.8\% of PSTW's Norwegian records also appeared in Norway's country-specific inventory, PSTW contained counterparts for only 26.8\% of the larger Norwegian inventory. France exhibited more modest overlap with PSTW: its 17 verified matches produced a Jaccard overlap of $0.093$ and represented 14.2\% of the French inventory and 21.3\% of PSTW's France records. Overlap was substantially lower for the Netherlands and the UK. PIL/USAID exhibited almost no overlap with the corresponding country-specific inventories: only one of its 10 US records matched the federal inventory, and no matches were verified for Mexico, the UK, or Uruguay. The latter country subsets contained only two to four PIL/USAID records, so these null overlaps should be interpreted in light of their limited size. Table~\ref{tab:cross-source-overlap} reports the results for every source pair.

\begin{table}[!ht]
    \centering
    \footnotesize
    \setlength{\tabcolsep}{2pt}
    \renewcommand{\arraystretch}{1.08}
    \caption{Overlap between country-specific and transnational sources}
    \label{tab:cross-source-overlap}
    
    \begin{tabularx}{\columnwidth}{
        @{}>{\raggedright\arraybackslash}X|l|r|r|r|r@{}
    }
        \toprule
        Country-specific & Transnational source &
        \shortstack{Records\\(C/T)} &
        Shared &
        Jaccard &
        \shortstack{Coverage (\%)\\(C/T)} \\
        \midrule
        France         & PSTW       & 120/80   & 17 & 0.093  & 14.2/21.3 \\
        Mexico         & PIL/USAID  & 119/4    & 0  & 0.000  & 0.0/0.0 \\
        Netherlands    & PSTW       & 1507/168& 9  & 0.005  & 0.6/5.4 \\
        Norway         & PSTW       & 179/65   & 48 & 0.245  & 26.8/73.8 \\
        UK & PIL/USAID  & 133/3    & 0  & 0.000  & 0.0/0.0 \\
        UK & PSTW       & 133/135  & 6  & 0.023  & 4.5/4.4 \\
        US  & PIL/USAID  & 3611/10 & 1  & 0.0003 & 0.03/10.0 \\
        Uruguay        & PIL/USAID  & 28/2     & 0  & 0.000  & 0.0/0.0 \\
        \bottomrule
    \end{tabularx}

    \vspace{0.4em}
    \begin{minipage}{\columnwidth}
        \emph{Notes:} C/T denotes country-specific/transnational.
        Coverage is the percentage of records in each source with a
        verified counterpart. Jaccard overlap is the number of shared
        records divided by the total distinct records across the source pair.
    \end{minipage}
\end{table}

Shared records were not consistently more complete than source-unique records. Within the French inventory, shared records had a mean disclosure-completion rate of 87.1\%, compared with 82.3\% among source-unique records. A similar difference appeared in Norway (70.8\% vs 62.3\%). In the Netherlands, however, shared records were less complete than source-unique records (77.8\% vs 84.3\%), while the difference in the UK was negligible (100.0\% vs 99.6\%). On the PSTW side of these comparisons, shared and source-unique records generally differed by no more than two percentage points.

Lifecycle composition also differed in some source pairs. In the Netherlands register, six of the nine shared records (66.7\%) were post-development, compared with 3.8\% of source-unique records; 91.5\% of source-unique Dutch records were operational. Within PSTW's Norway subset, 68.8\% of shared records were classified as pilots, compared with 35.3\% of source-unique records. Overall, country-specific and transnational sources intersected selectively rather than reproducing the same portfolios.
\section{Discussion}\label{sec:discussion}
While most AI registers shared a descriptive core, they differed in scope, schema, field completion, and the organizations and systems represented. Schema breadth did not necessarily produce greater transparency: broad schemas often contained substantial missingness, whereas narrower schemas could achieve higher completion by requesting less information. The lack of strong evidence for temporal or regional groupings in the hierarchical ordering complicates straightforward accounts of policy diffusion. As local administrative settings translate and adapt policy scope, terminology, and implementation~\cite{marsh2013policy}, schema similarity alone cannot establish borrowing, and the diffusion of a common policy form, such as register schema, need not produce convergence in operational design. Our cross-source comparisons further demonstrate that visibility is distributed across a source ecology, as country-specific and transnational sources frequently disclosed different systems within the same jurisdiction. Low overlap between such sources may reflect different definitions or institutional reach rather than inaccuracy. Here, neither source type should be treated automatically as ground truth, but together be considered as evidence of representational complementarity. In such cases, rather than simply aggregating AI registers, triangulation can expand visibility when provenance and source-specific definitions are preserved. Hence, AI registers should be understood not as censuses, but as institutionally produced and complementary representations of public-sector AI. Drawing on infrastructure and policy instrumentation scholarship, we propose the \emph{\textbf{Layered Visibility Framework}}, which conceptualizes how this representation is produced through four interconnected layers.


The first layer, \emph{institutional scope}, determines which entities and activities enter a register. Since reporting entities may decide differently to include pilot or retired systems and use mandatory, voluntary, centrally compiled, or independently assembled reporting arrangements, differences in reporting mandates, participation, institutional coverage, collection model, or units of observation may shape organizational concentration and lifecycle composition independent of underlying patterns of governmental AI.

The second layer, \emph{schema affordances}--the possibilities for disclosure enabled and constrained by its design~\cite{evans2017explicating}--determines what an instrument can represent. Most schemas included names, responsible organizations, descriptions, technology types, and lifecycle statuses, suggesting a shared descriptive core. In contrast, only a few registers' schemas explicitly requested information about risks, legal bases, external evaluations, and appeal processes. Because information absent from a schema cannot be disclosed systematically, schema design establishes the boundaries of accountable knowledge~\cite{poirier2021reading, kuehnert2025and}. The observed variation in the number of fields shows that instruments identically labeled as AI registers may enable substantially different forms of scrutiny.

The third layer, \emph{reporting realization}, concerns whether these possibilities are realized in individual records. Schema presence and record-level disclosure were not equivalent. For example, the US had the broadest schema but many fields (e.g., appeal process) had a low completion rate, whereas several narrower instruments fully populated their eligible fields. Thus, registers can be broad but thin or narrow but dense, while the expectation is that they be broad and consistently populated. Counting fields alone overstates the visibility afforded by broad but incomplete schemas, while completion rates alone favor schemas that request less information.

These layers produce the fourth: the \emph{public representation} of governmental AI. Lifecycle composition and organizational concentration show that register records constitute portfolios produced through particular disclosure arrangements. Selective cross-source overlap also demonstrates that different instruments can produce different representations of the same jurisdiction. Hence, register records are not unmediated representations of governmental AI.

The framework extends the evaluation of registers beyond transparency to representational quality and interoperability. Instead of imposing a universal template, registers could combine a clearly defined core--such as system ID, responsible organization, lifecycle stage, and last-updated date--with modules covering procurement, data use, human involvement, risk, evaluation, and contestability. Distinguishing structural absence from missing reporting, publishing category definitions, describing coordination processes, documenting schema changes, and preserving source links would support comparison without erasing institutional differences.
\section{Future Work and Conclusion}\label{sec:conclusion}
AI registers are becoming part of the administrative infrastructure through which governments render digital transformation knowable and governable. However, similarly labeled instruments can produce distinct and potentially complementary representations of governmental AI. Besides being necessarily partial, the publicly accessible instruments we studied capture particular points in time, which limited our record linkage capacity and our ability to map source-specific vocabularies. Future work should track how registers and schemas change over time to accommodate mappings for heterogeneous categories while preserving corresponding institutional meanings. Interoperability among these instruments depends on shared concepts, clear definitions, preserved provenance, and traceable differences for accountable digital transformation.

\begin{acks}
We used Codex for debugging and Grammarly for language editing.
\end{acks}

\bibliographystyle{ACM-Reference-Format}
\bibliography{sample-base}


\end{document}